\documentclass[times, 10pt,twocolumn]{article} 
\usepackage{latex8}
\usepackage{times}

\usepackage{graphics} % for pdf, bitmapped graphics files
\usepackage{epsfig} % for postscript graphics files

\begin{document}

\title{  $\lambda$-Connectedness Determination for Image Segmentation}

\author{Li Chen  \\
    Department of  Computer Science 
        and Information Technology\\
    University of the District of Columbia \\
      Washington, D.C. 20008, USA \\
       lchen@udc.edu\\ 
\noindent{\small \it 36th Applied Image Pattern Recognition Workshop (AIPR 2007), October 2007, Washington, DC, USA, IEEE Proceedings.}\\
}

\maketitle

\begin {abstract}

Image segmentation is to separate an image into distinct homogeneous regions 
belonging
to different objects. It is an essential step in image analysis and computer 
vision.  
This paper compares some segmentation technologies and attempts to find 
an automated way to better determine the parameters for image segmentation, 
especially the connectivity value of
$\lambda$ in $\lambda$-connected segmentation.

Based on the theories of the maximum entropy method and Otsu's minimum 
variance method, 
we propose:(1)maximum entropy connectedness determination: a method that 
uses maximum entropy to 
determine the best $\lambda$ value in $\lambda$-connected segmentation, and 
(2) minimum variance connectedness determination:  
a method that uses the principle of minimum variance to determine $\lambda$ value.
Applying these optimization techniques in real images,  
the experimental results have shown great promise in the development 
of the new methods. 
In the end, we extend the above method to more general case in order 
to compare it with the
famous Mumford-Shah method that uses variational principle and geometric measure.

\end {abstract}

%%%%%%%%%%%%%%%%%%%%%%%%%%%%%%%%%%%%%%
%%%%%%%

\section {Introduction}
 
Image segmentation is the basic approach in image processing and computer vision \cite{PP}.
It is used to locate special regions and then extract information from them. 
Image segmentation is used to partition an image into different
components or objects and is an essential procedure for image preprocessing, 
object detection and extraction, and object tracking. Image segmentation 
is also related to edge detection. 

Even though there is no unified theory for image segmentation , 
some practical methods have been studied over the years such as 
thresholding, edge based segmentation, region growing, clustering 
(unsupervised classification), and split-and-merge segmentation, to name a few. 
 $\lambda$-connected segmentation is a technique in 
 the category of region growing segmentation. 
 It was proposed to find an object having the property of gradual variation \cite{Che85}
 \cite{CCZ} \cite{Che91} \cite{CA} \cite{CAC}.
 
%%%%%%%%%%%%%%%%%%%%%%%
%%%%%%%%%%%
This paper attempts to find 
an automated way to better determine the parameters for image segmentation, especially the connectivity value of
$\lambda$ in $\lambda$-connected segmentation.

This paper first reviews some major segmentation techniques to explain why segmentation is difficult,
and how a special technique would be selected in specific applications. 
We then focus on our problem of determining
segmentation parameters in $\lambda$-connected segmentation.
Based on the philosophies of the maximum entropy method and Otsu's minimum variance method, 
we propose: (1)maximum entropy connectedness determination: a method that uses maximum entropy to 
determine $\lambda$ value, and (2) minimum variance connectedness determination:  
a method that uses the principle of minimum variance to determine $\lambda$ value.
Applying these optimization techniques in real images,  
the experimental results have shown the great promise of the new methods. 
In the end, we extend the above method to a more general case in order to compare it with the
famous Mumford-Shah method that uses variational principle and geometric measure \cite{MS}.

%%%%%%%%%%%%%%%%%%%%%%%%%%%%%%%%%%%%%%%%%%%%%%%%%%%%%
%%%%%%%%%%%

\section { Image Segmentation Review } 

In this section, we first review currect technology of image segmentation: five types of 
techniques, their characteristics, and uses. We then focus on the connectedness-based image
segmentation technique.  

\subsection { Overview of Image Segmentation Approaches}

As we know, there is no unified theory for image segmentation, 
 some practical methods have been studied over the years such as thresholding, 
 edge based segmentation, region growing, clustering (unsupervised classification, e.g. k-mean or fuzzy c-mean), 
 and split-and-merge segmentation.  These segmentation algorithms have been developed for
solving different problems \cite{SW}.  However, they are all based on one or more of the five philosophies listed below:

(1) A segment is a class/cluster, so one
can use a classification/clustering method to segment the image. Classification methods usually do not
need to use the location/position information. Clustering for unsupervised classification technology can
perform better  to find an object for sampled points within  the subset of data frames.
Typical techniques include Isodata and  k-mean or fuzzy c-mean.

  $K$-mean or fuzzy $c$-mean is a standard classification method that is
often used in image segmentation \cite{Hal}.
This method classifies the pixels into different groups in order to minimize the
 total ``errors,'' where the ``error'' is  the distance from the pixel value 
 to the center of its own group.

(2) A segment is a homogeneous region. If an object or region can be identified 
by absolute intensity (the pixel value), we usually use threshold segmentation. 
In other words, an object will be recognized as a geometrically 
connected region whose values/intensities are between a certain high-limit and a low-limit. 
We usually assume that the high limit is the highest value of the image. Therefore,
in practice, one only needs 
to determine the low-limit. Maximum entropy and minimum variance (also called
Otsu's method) are two 
of the most popular methods for determining the best threshold for single image \cite{STI}.  

Multilevel thresholding is similar to threshold segmentation and uses the same
philosophy, but multiple thresholds are produced at once. It needs an extremely high 
 time cost for computation \cite{PP} \cite{WH}. We will discuss these two methods in detail in section 3.

(3) A segment is a ``smoothly-connected'' region.
In a region where intensity changes smoothly or gradually, the region is viewed as a segment.
 Smoothness can be measured by a limit. 

A popular segmentation method is called mean-based region growing
segmentation in this paper \cite{GW}.
 A pixel will be included in a region if the updated
 region is homogeneous, meaning that the difference between
 pixel intensity and the mean of the region is limited by $\epsilon$, a small 
 real number. 

The $\lambda$-connected segmentation follows the same philosophy. This is to link all pixels 
that have the similar intensity.

This method is related to a fuzzy method created by Rosenfeld who treated an image as a 2D fuzzy set.
Then, he used $\alpha$-cut to segment the image into components. Another way is to measure two pixels 
to see if they are ``fuzzy'' connected.
A pixel set is $\lambda$-connected if for any two points there
is a path that is $\lambda$-connected where $\lambda$ is a fuzzy value
between 0 and 1 \cite{CCZ}\cite{Che85}\cite{Che91} \cite{TB}\cite{TBL}. This is a generalization of threshold
segmentation
in some cases.
This method can be used to divide (partition) 
different intensity levels without calculating different thresholds or clip-level
values. However, for a complex image,
how to calculate the value of $\lambda$ remains unknown. 

A fast algorithm can be designed to perform a
segmentation. In fact, the simple form of both threshold segmentation and $\lambda$-connected segmentation can be
done in linear time.

%%%%%%%%%%%%%%%

We can see that mean-based region growing segmentation is a statistical approach, 
but the $\lambda$-connected approach is a graph-theoretic method. We can combine them by 
requiring that the 
$\lambda$-connected segment also be within a $\epsilon$ limit of the mean. 
Or after the mean-based segmentation, we can do a $\lambda$-connected segmentation.\cite{Che06}

(4)Split-and-merge segmentation uses quadtree to determine the order in which pixel(s) 
should be treated or computed \cite{Pav}. It is an algorithmic way to find an object or to force a
merge order. This is because this method is based on the mean of the merged region. It does not
guarrantee a transitive relation. Again, the mean-based segmentation is not
an equivalence relation. 
This method splits an image into four sections and
checks if each part is homogenous. The homogenous segments are then merged together. 
If the segments are not homogenous, the splitting process is repeated.
This process is
also called quadtree segmentation. The method is more accurate
for some complex images.
However, it costs more time to segment an image.
The time complexity of this method (process) is  $O(n log n )$. 
This was proved by Chen in 1991.\cite{Che91}

(5) A segment is surrounded by one or several closed edges.
If we can detect and track the
edges, we can determine the location and outline of the segments.

The fifth philosophy is edge detection. To find low or high frequency pixels are very common in edge detection. 
However, not all edge-detection methods can be used in image segmentation since enhancing edge is not the primary purpose
of image segmentation. The purpose of image segmentation 
is to find components.
 The number of edges should be relatively small. Otherwise, the extraction of the closed curves will 
be the major problem. Recent development indicated that the Mumford-Shah method is promising.  The method uses
the variational principal \cite{MS}. This method has captured
a considerable amount of attention.

The Mumford-Shah method considers three factors in segmentation:
(1) the total length of all the segments?? edges, (2) the
unevenness of the image without its edges, and (3) the total error between the original
image and the
proposed segmented images where each segment has unique or similar values in its pixels.
When the three weighted   factors are minimized,
the resulting image is a solution of
the Mumford-Shah method. Recently, Zhan and Vese proposed level-sets to
simplify the Mumford-Shah method so that it
produced better results \cite{CV}. However, level-sets use contour boundaries
that may limit the flexibility of the original
Mumford-Shah method. The Mumford-Shah method also needs an alternative process and
its algorithm performance is still unknown.

\subsection{Remarks on Selecting an Appropriate Segmentation Method}

The $k$-mean or fuzzy c-mean, maximum entropy, and the Mumford-Shah method all
require an iterated process
that is good at  detailed or fine segmentation. This is not a quick solution
for fast segmentation.
The fast segmentation methods are only used for region growing including the original
threshold method, mean-based and lambda-connected search, and split-and-merge method.

In practice, $k$-mean or fuzzy c-mean and  maximum entropy are still the most popular. However, 
for some types of sequential images, these two methods do not obtain stable results
when  we process a set of meteorological data to find areas with the most water vapor
indicating (most likely) the location of a hurricane \cite{LKZC}. $\lambda$-connected segmentation, 
on the other hand, has worked very well. A problem the $\lambda$-connected method faced 
was that the $\lambda$ value needed to be assigned even though that value can be used 
throughout each data frame in the data set. To find a way to automatically 
  determine the $\lambda$-value is a longtime goal for  the author. 
  Some methods have been proposed such as maximum connectedness spanning tree \cite{Che02} \cite{Che04} 
  and the golden cut point \cite{Che06}. 

  In this paper, the author proposes  optimization methods based on maximum entropy 
and minimum variance, respectively.

\subsection {The Connectedness-based Segmentation}

$\lambda$-Connected segmentation is based on the philosophy that an object must
Have a smooth inside and has a gap at its boundary (in terms of intensity).
Trying to find a closed curve/boundary that indicates the intensity of the
jump is the key to this method.

A measure of connectedness can be used to partition a set of data
into connected components
based on adjacency or neighborhood systems.
Using connectedness to divide an image does not require transforming the
image into binary form.  After the data is partitioned, a fast algorithm  such as
the breadth-first-search algorithm 
can be applied to find a connected component \cite{Che85}\cite{CCZ}.
$\lambda$-connected
search was introduced to segment such an image without transferring
the image into a $\{0,1\}$-image.  However, the value of $\lambda$, usually between
0 and 1, determines the fineness of the segmentation.

$\lambda$-connectedness can be defined on an undirected graph $G=(V,E)$
with an associated (potential) function $f:V\rightarrow R^{m}$, where $R^{m}$ is
the $m$-dimensional real space \cite{CAC}. Given a measure
$\alpha_{\rho}(x,y)$ on each
pair of adjacent points $x,y$
based on the values $\rho(x),\rho(y)$,   we define
\setlength{\arraycolsep}{0pt}
% changing \arraycolsep reduces excess space for wrap-around
% equation
\begin{eqnarray}
  \alpha_{\rho}(x,y) = \left\{\begin{array}{ll}
              \mu(\rho(x),\rho(y)) & \mbox{ if $x$ and $y$ are adjacent}\\
              0            &   \mbox{ otherwise}
              \end{array}
              \right.
\end{eqnarray}
\noindent where $\mu: R^{m}\times R^{m} \rightarrow [0,1]$ with
$\mu(u,v)=\mu(v,u)$ and $\mu(u,u)=1$.
$\alpha_{\rho}$ is used to measure ``neighbor-connectivity.'' The next step is to
develop path-connectivity so that
$\lambda$-connectedness on $<G,\rho>$ can be defined in a general
way.

 In graph theory,	a finite sequence $x_1,x_2,...,x_n$ is
called a path, if $(x_i,x_{i+1})\in E$.
The path-connectivity $\beta$ of a path $\pi=\pi(x_1,x_n)=\{x_1,x_2,...,x_n\}$
is defined as
\setlength{\arraycolsep}{0pt}
\begin{eqnarray}
 \beta_{\rho}(\pi(x_1,x_n)) =  \min \{ \alpha_{\rho}(x_i,x_{i+1}) |
i=1,...,n-1 \}
\end{eqnarray}
\noindent or
\setlength{\arraycolsep}{0pt}
\begin{eqnarray}
 \beta_{\rho}(\pi(x_1,x_n)) =   \prod  \{ \alpha_{\rho}(x_i,x_{i+1}) |
i=1,...,n-1 \}
\end{eqnarray}
\noindent Finally,  the degree of connectedness or connectivity of two
vertices $x,y$ with respect to $\rho$ is defined as:
\setlength{\arraycolsep}{0pt}
\begin{eqnarray}
 C_{\rho}(x,y) = \max \{ \beta(\pi(x,y)) | \pi \mbox { is a (simple) path}. \}
\end{eqnarray}
For a given $\lambda \in [0,1]$, point $p=(x,\rho(x))$ and $q=(y,\rho(y))$
are said to be $\lambda$-connected
if $C_{\rho}(x,y)\ge \lambda$.
In image processing,
$\rho(x)$ is the intensity of a point $x$ and   $p=(x,\rho(x))$ defines a pixel.

%%%%%%%%%%%%%%%%%%%%%%%%%%%%%%%%%%%%%%%
%%%%%%%%

\section {$\lambda$ Value Determination and Optimization}

It is a natural and unavoidable question how we determine $\lambda$ value in connectedness-based 
segmentation? It is somehow similar to determine the clip level in threshold segmentation; however, 
$lambda$ value is not as sensitive as the clip-level in thresholding. There are fewer $\lambda$ values to be
selected than clip-levels. In Chapter 10 of \cite{Che04}, Chen provided the detail analysis on this issue.  

Some techniques have been proposed and tested such as the binary search-based method and the maximum 
connectedness spanning tree method \cite{Che04} \cite{Che02}.  We also proposed a golden-cut 
technique for bone density measurement in \cite{Che06}.
In this section, we propose two new methods to determine the $\lambda$ value 
for the segmentation. The new methods are based on maximum entropy 
and minimum variance, respectively.

\subsection {Connectedness and Maximum Entropy} 

In this subsection, we propose a method that uses the maximum entropy method to determine  
the $\lambda$ value. It can be called maximum entropy connectedness 
determination. 

The maximum entropy method was first proposed by Kapur, Sahoo, and Wong $\cite{KSW}$.  
It is based on the maximization of inner entropy in both the foreground  and background. 
The purpose of finding the best threshold is to make both objects in the 
foreground  and background, respectively, as smooth
as possible. \cite {KSW} \cite {PP} \cite{STI}

If  $F$ and $B$ are in the foreground and background classes, 
respectively, the maximum entropy can be calculated as follows; 

\[ H_{F}(t)=  -\Sigma_{i=0}^{t} \frac{p_{i}}{p(F)} \ln \frac{p_{i}}{p(F)} \]

\[ H_{B}(t)=  -\Sigma_{i=t+1}^{255} \frac{p_{i}}{p(B)} \ln \frac{p_{i}}{p(B)} \]

\noindent where $p_{i}$ can be viewed as the number of pixels whose value is $i$; 
$p(B)$ is the number of pixels in background, and 
$p(F)$ is the number of pixels in foreground. 
The maximum entropy is to find the threshold value $t$ 
that maximizes $H_{F}(t) + H_{B}(t)$.

Such an idea can be used for $\lambda$-connected segmentation.
However, the total inner entropy for the image is to calculate the entropy for each 
segment ($\lambda$-connected component), not for the thresholding clipped 
foreground/background.  This is because in $\lambda$-connected segmentation
there is no specific background. Each $\lambda$-connected segment can be viewed as
foreground, and the rest may be viewed as the background. 
It is different from the original maximum entropy where the
range of pixel values determines the inclusion of pixels.
Therefore, we need to summarize all inner entropies in all segments.

\begin{equation}
          H (\lambda ) = \Sigma (\mbox{inner entropy of each $\lambda$-connected component}) 
\end{equation}

We will select the $\lambda$ such that $H(\lambda)$ will be maximized.
We call this $\lambda$-value the maximum entropy connectedness. This unique value is
a new measure for images.

Since the maximum entropy means the minimum amount of information or minimum variation, 
 we want the minimum change inside each segment. This matches the philosophy  of
the original maximum entropy method.  In other words, the $\lambda$-connected  maximum entropy  has a better meaning in
some applications.  We  use the $\lambda_e$ such that 

\[H(\lambda_e) = \max \{ H(x) | x\in [0,1] \}.\]

Assume there are $m$ $\lambda$-components, define inner entropy of each 
$\lambda$-component $S_{i}$:

\[ H_i(\lambda)= \Sigma_{k=0}^{255}- \frac{Histogram[k]}{n} \log \frac {Histogram[k]} {n} \]

\noindent where $n$ is the number of points in the component $S_i$. $Histogram[k]$ is
the number of pixels whose values are $k$ in the segment. Thus,

\begin{equation}
          H (\lambda ) = \Sigma_{i=1}^{m} H_i(\lambda)) 
\end{equation}

The maximum entropy connectedness can be viewed as a measure
of a special connectivity  for the image. If $\lambda$ value 
is calculated in the above formula for an image that makes 
$H(\lambda)$ to be maximum, we call that the
image have the maximum entropy connectedness $\lambda$, denoted as
$\lambda_e$.

\subsection{Experimental results with  $\lambda_e$}

In \cite{Che06}, we proposed a golden cut method for finding the $\lambda$-value for bone
density connectedness calculation. We have obtained a $\lambda$=0.96, 0.97 for a bone image 
(the size of the picture is different from the one used in this paper). 
For a similar image, using the maximum entropy connectedness presented in this section, we got
$\lambda_e$=0.95. The result is quite reasonable. The original image and 
both of the segmented images are shown in Fig. 1-4. No pre-cut (preprocessing) 
is performed in the segmentation.

%%%%%%%%%%%%%%%%%%%%%Fig. 7.9.
\begin{figure}[hbt]
	\begin{center}

   \epsfxsize=1in 
   \epsfbox{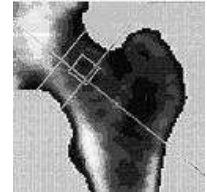}

	\end{center}
\caption{ Bone Density Image Segmentation : the Original image}
\end{figure}

%%%%%%%%%%%%%%%%%%%%%Fig. 7.9.
\begin{figure}[hbt]
	\begin{center}

   \epsfxsize=1in 
   \epsfbox{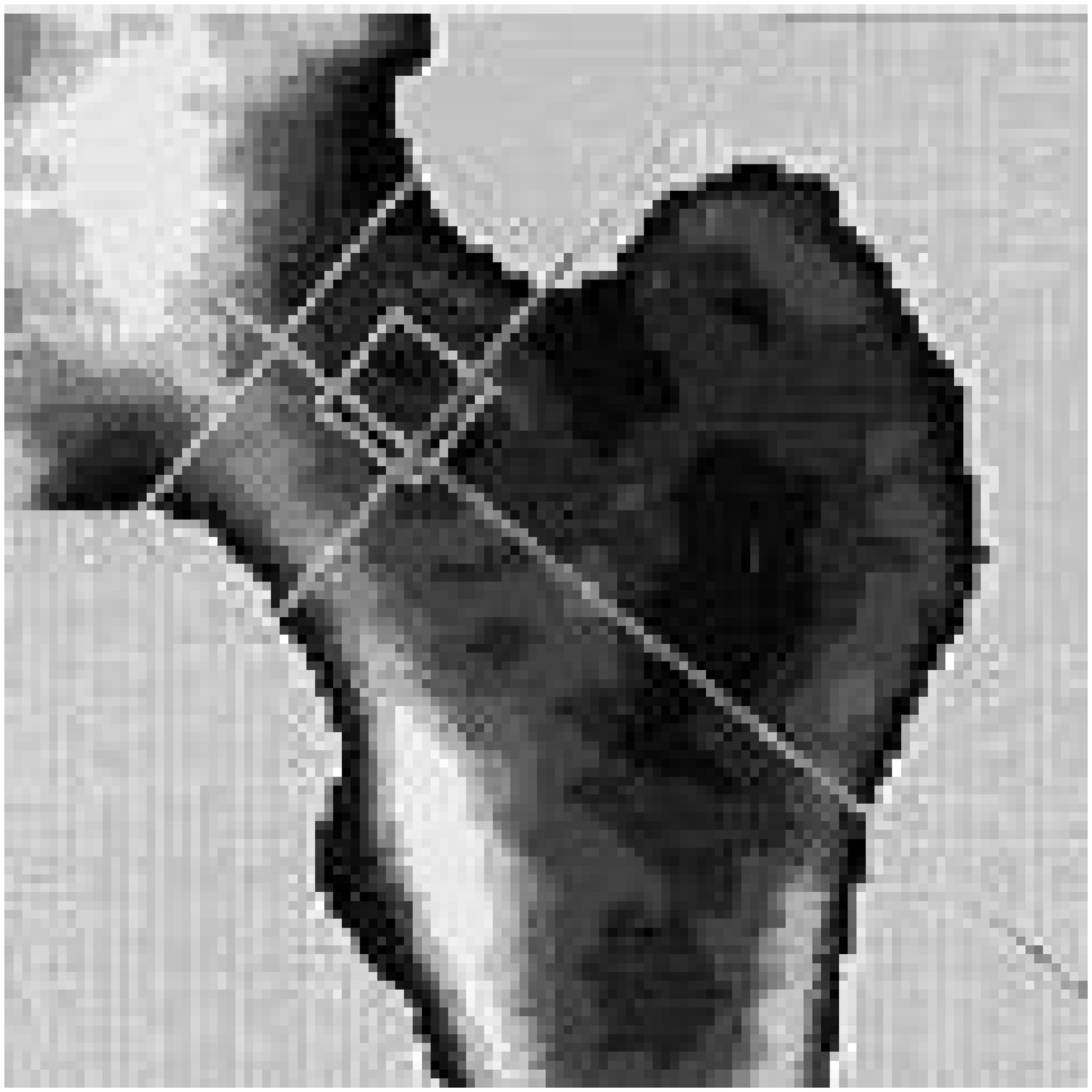}

	\end{center}
\caption{ Bone Density Image Segmentation : $\lambda$=0.97}
\end{figure}

%%%%%%%%%%%%%%%%%%%%%Fig. 7.9.
\begin{figure}[hbt]
	\begin{center}

   \epsfxsize=1in 
   \epsfbox{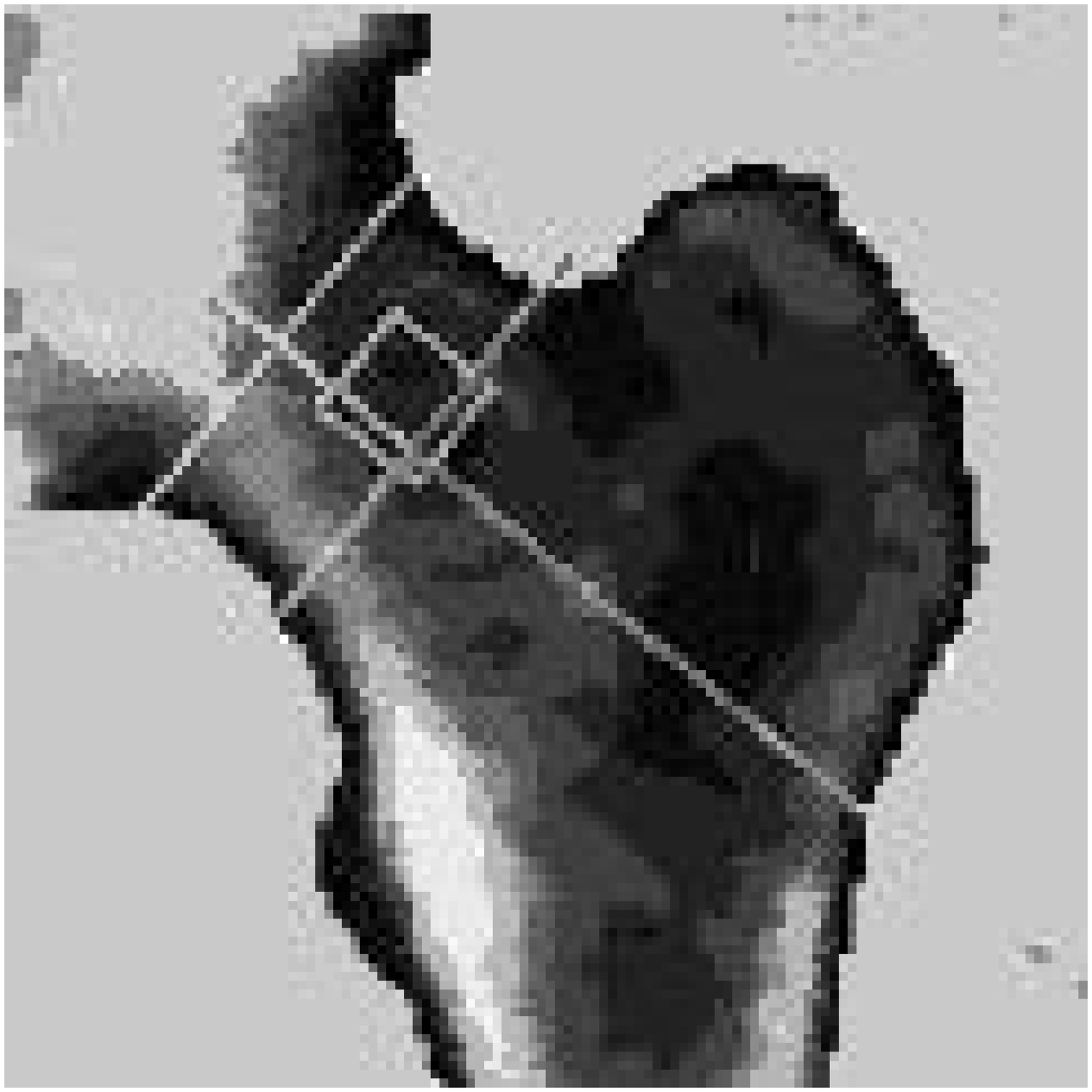}

	\end{center}
\caption{ Bone Density Image Segmentation : $\lambda_e$=0.95}
\end{figure}

%%%%%%%%%%%%%%%%%%%%%Fig. 7.9.
\begin{figure}[hbt]
	\begin{center}

   \epsfxsize=1in 
   \epsfbox{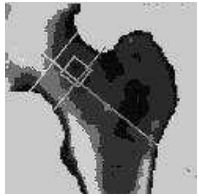}

	\end{center}
\caption{ Bone Density Image Segmentation : $\lambda$=0.93}
\end{figure}

What we can see in the above three segmented images (Fig. 2-4). Fig. 2 seemed to be
the same as the original image Fig. 1. It is possible that Fig. 4 may represent the
better understanding of bone connectivity. However, what we state here is that
$\lambda_e$ can provide us the meaningful result and it was done automatically.

For the commonly used testing image "Lena," Fig.5, the result of $\lambda$-connected segmentation 
is quite interesting. Whatever 
we use a pre-threshold cut or not, $\lambda_e$ is always 0.99 (Fig. 6). An original maximum
entropy arrived at the clip-level of 125 counts of the 8-bit gray level image (0-255 
of the pixel value range. The reason is that the ``Lena'' image does not contain many
``continuous'' parts. In the $\lambda$-connected segmentation, we can see
that $\lambda_e$ (=0.99) connected segmentation has connected the continuous component especially 
at the face and shoulder. This matches the result of using the maximum entropy cut, Fig. 7 (We use 
NIH $ImageJ$ to perform the cut.) Thus, we can say that our new method is still reasonable. 
When we use $\lambda =0.98$ for the image, we get Fig. 8.

%%%%%%%%%%%%%%%%%%%%%Fig. 7.9.
\begin{figure}[hbt]
	\begin{center}

   \epsfxsize=2in 
   \epsfbox{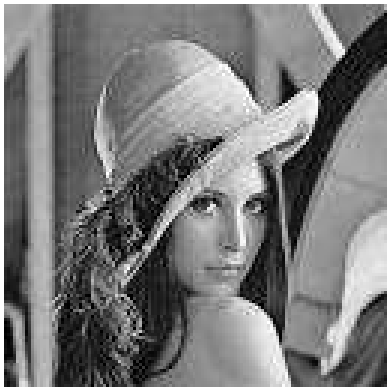}

	\end{center}
\caption{  Image Segmentation for testing image "Lena:" the Original image }
\end{figure}

%%%%%%%%%%%%%%%%%%%%%Fig. 7.9.
\begin{figure}[hbt]
	\begin{center}

   \epsfxsize=2in 
   \epsfbox{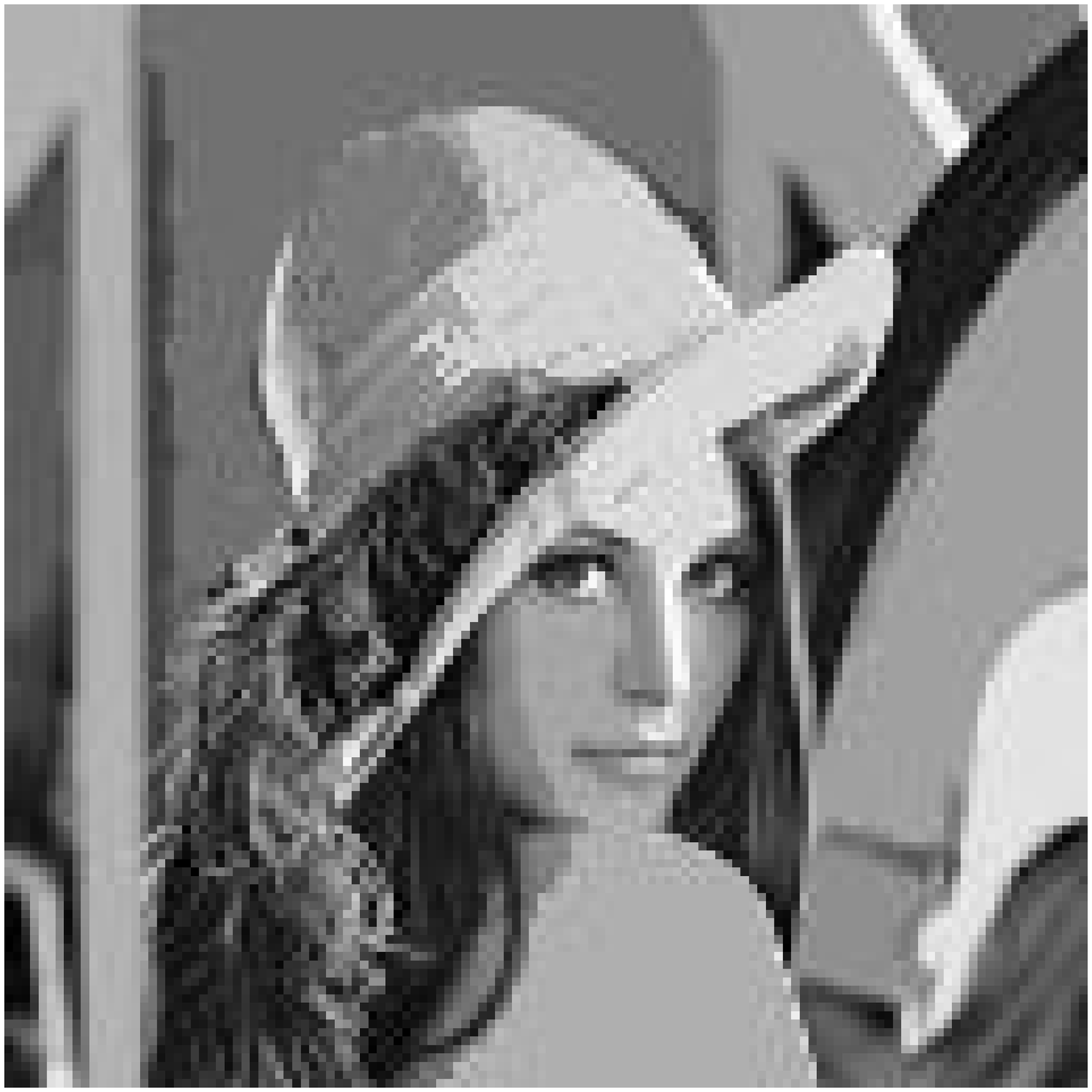}

	\end{center}
\caption{  Image Segmentation for testing image "Lena:" $\lambda$=0.99  }
\end{figure}

%%%%%%%%%%%%%%%%%%%%%Fig. 7.9.
\begin{figure}[hbt]
	\begin{center}

   \epsfxsize=2in 
   \epsfbox{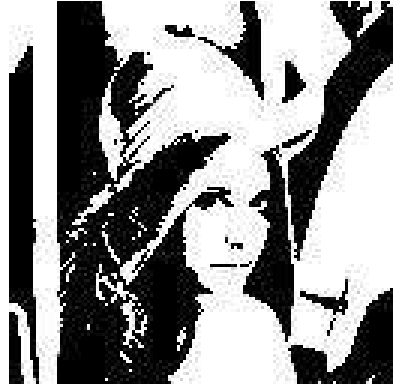}

	\end{center}
\caption{  Image Segmentation for testing image "Lena:" Standard Maximum Extropy }
\end{figure}

%%%%%%%%%%%%%%%%%%%%%Fig. 7.9.
\begin{figure}[hbt]
	\begin{center}

   \epsfxsize=2in 
   \epsfbox{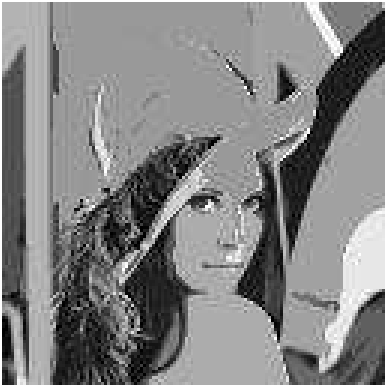}

	\end{center}
\caption{  Image Segmentation for testing image "Lena:" $\lambda$=0.98  }
\end{figure}

For the image having gradual variation property, 
$\lambda$-connected segmentation usually has an advantage.
We have extensively tested a set of sequential images in order to find the outlier
of meteorological data that  indicates (most likely) the hurricane center 
\cite{LKZC}. The data frames we used are  water vapor images. 

Except the standard threshold method and the $\lambda$-connected segmentation method, 
all other methods we tested failed including famous $k$-mean and maximum entropy \cite{LKZC}. What 
we found was that the key for this sequential images is that the pre-cut is necessary. 
Interesting enough for us,  $45\%$ of the pick value of each image for the cut receives the best
result. After that, we can use $\lambda$=0.95 for the segmentation parameter and extract
the largest component for the outlier searching result.  

The problem here is that $45\%$ of the pick value as clip-level is not automatically calculated.
If we want a totally automatic process,  
we might to use maximum entropy to make the first cut. 
In this testing set, we have 12 image frames. For some beginning images, 
the method worked well as expected.  For other images, the new method proposed in this paper 
consistently got wrong results since 
the $\lambda_e$ calculated is always greater than or equal to 0.97. We cannot get $\lambda$ to be 
0.95 using ``maximum entropy connectedness.'' 
After looking into the detail images, we have found two problems: (1) the pixel values 
are not ``continuous'' around ``the desired outlier,'' and (2) The largest component
criteria for the outlier is not quite represent the nature of hurricane centers 
(outliers), we need to change the criteria to ``the 
largest and the brightest.'' 

For (1), we have done a smoothing process. For (2), we change the outlier criteria from the largest component
 to the total
intensity of the component (not only testing its size). Smoothing preprocess is reasonable, and it
is good for the $\lambda$-connected segmentation to find large component.    

The following image shows such a treatment. We have applied an automatic pre-cut by using
maximum entropy instead of the pre-cut using the $45\%$ of pick value. Fig. 9 shows an original
image.

%%%%%%%%%%%%%%%%%%%%%Fig. 7.9.
\begin{figure}[hbt]
	\begin{center}

   \epsfxsize=3in 
   \epsfbox{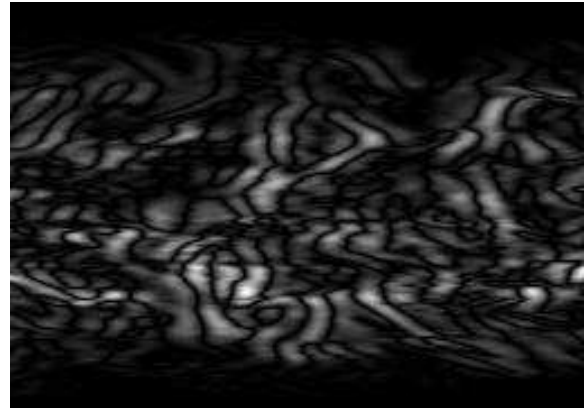}

	\end{center}
\caption{ Image Segmentation for Meteorological data outlier: Original Image }
\end{figure}

Fig. 10 shows the result using our new method plus
a maximum entropy threshold cut of preprocessing. 
A threshold value=23 is calculated by standard maximum entropy. 
Then, we perform the automated finding process to get $\lambda_e$ described above. 
$\lambda_e=0.90$ was obtained and used.  

\begin{figure}[hbt]
	\begin{center}

   \epsfxsize=3in 
   \epsfbox{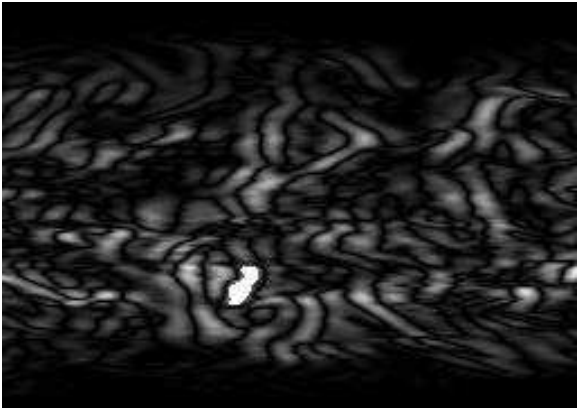}

	\end{center}
\caption{Image Segmentation for Meteorological data outlier: $\lambda_e=0.90$,
 maximum entropy connectedness determination. The image is smoothed and pre-cut
 by the standard maximum entropy at threshold = 23.}
\end{figure}

If we just use the standard maximum entropy at threshold = 23, we will have the 
following image Fig. 11.  

\begin{figure}[hbt]
	\begin{center}

   \epsfxsize=3in 
   \epsfbox{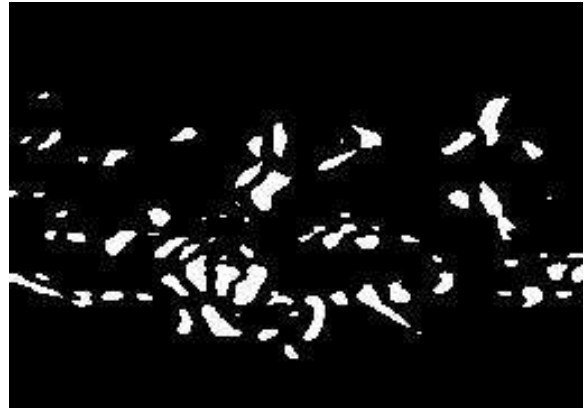}

	\end{center}
\caption{Image Segmentation for Meteorological data outlier:  The image is smoothed 
  and clipped by standard maximum entropy at threshold = 23.}
\end{figure}

\subsection{Consider Outer Entropy in Maximum Entropy Connectedness }

Using maximum entropy is a type of philosophical change. In fact, we can
consider other formulas. For example we can select 
another way to calculate the  
entropy of one segment. 

Here we propose a different formula. We can calculate the inner entropy of a component, 
then treat the rest of the image as the 
background for the component. The total entropy generated by this segment is
the summation of both. 
We can apply this process to all components/segments while segmenting.  

 Let $I$ be the image, $C_i(\lambda) = I-S_i(\lambda)$ is the complement of component $S_i(\lambda)$

\[H(C_i(\lambda))= \{ \mbox{Entropy for the set $C_i(\lambda)$} \} \]

\noindent We can use the following formula for the the basis of optimzation.
\[H(\lambda) =  H(S_i(\lambda)) +  H(C_i(\lambda))\]

The outer (background) entropy is the total. 

\[H(outer)= \Sigma H(C_i) \]
 
\noindent In $H(outer)$, a pixel is calculated multiple times. 
We may need to use the average $H(outer)/m$ where $m$ is the number of segments. 
The relationship between this formula and the formula we used in the previous subsection
is also interesting.

Furthermore,  we should consider the following general model.

\[H_optimal = a \cdot H(inner) + b\cdot H(outer)\] 

\noindent where $a$ and $b$ can be constant or function of segmentations.

\subsection {Connectedness and Minimum (Inner) Variance}

In this subsection, we develop a minimum variance-based method for finding 
the best $\lambda$ value in $\lambda$-connected segmentation. Minimum variance 
was first studied by 
Otsu in image segmentation $\cite{Ots}$ $\cite{CV}$. In other words, Otsu's 
segmentation was the first global optimization solution for
image segmentation. It is used to clip the image into two parts: the object
and the background.

Assume that $\sigma^{2}(W)$,$\sigma^{2}(B)$,  $\sigma^{2}(T)$
represent the within-class
variance, between-class variance, and the total variance, respectively.
The optimum threshold will be determined by maximizing one of the
following criterion with respect to threshold $t$ $\cite{Ots}$ $\cite{CV}$:

\[\frac{\sigma^{2}(B)} {\sigma^{2}(W)}, \frac{\sigma^{2}(B)} {\sigma^{2}(T)} ,
\frac{\sigma^{2}(T)} {\sigma^{2}(W)} \]

\noindent $\sigma$ is the standard derivation. Since $\sigma^{2}(T)$ is constant for a certain image, 
this segmentation process is to make 
between-class variance large and within-class variance small. Therefore, our task is to
make the within-class variance as small as possible.

The original design of Otsu's method is not  able to be implemented directly 
for $\lambda$-connected segmentation which is similar to the case of maximum entropy.
This is because there were only
two categories, the object and the background. For the first and second criteria, 
in $\lambda$-connected segmentation there are many components and
it would be very hard to find between-class variance.  We could consider the 
total between-class variance, by considering every pair of components. Or we could
consider  between-class variance for the components that  are the neighbors . The third
criterion seems likely to be valid, however, when we only consider the variance within a 
connected component, what will happen is $\sigma^{2}(W)=0$ if 
$\lambda=1$.  $\frac{\sigma^{2}(T)} {\sigma^{2}(W)}$ will be infinite and will 
always be the greatest value.

In order to make use of Otsu's philosophy, we modify the original formula by adding 
a term that is the number of segments or components, $M$. We try to minimize the following
formula:

\begin {equation}   
     H (\lambda ) = \Sigma (\mbox{inner variance of each} \lambda-component) + c\cdot M  
\end {equation} 
 
\noindent where $c$ is a constant. We could let $c=1$.  
 The calculation of the inner variance of a $\lambda$-component is to compute 
 the variance (square of standard deviation) of the pixels in the component.  

The following formula is to find minimum average variance (for each component).
 
\begin {equation} 
    H (\lambda ) = \Sigma (\mbox{inner variance of each} \lambda-component) / M   
\end {equation} 

We want to find $\lambda_v$ such that 
\begin {equation} 
    H (\lambda_v ) = \min \{ H (\lambda ) |    \lambda \in [0,1]\}
\end {equation}

This strategy only works for the meteorological data. 
The experimental results show that the method is promising.
 For the other two kinds of images tested in maximum entropy connectedness, 
 ``Lena'' and the Bone image,
we still need to find an appropriate way under minimum variance philosophy.

The following images show the process on the same picture with
a preprocessing threshold cut using the maximum entropy cut or $45\%$ peak cut.
Then, we perform the automated process of finding of $\lambda$-value.
Fig. 12 shows that we arrived at  $\lambda_v$=0.97 using the method 
of minimum variance connectedness determination
described in this subsection. Without  smoothing the original image, we pre cut the image
using maximum entropy threshold. The result is not what we expected. 

\begin{figure}[hbt]
	\begin{center}

   \epsfxsize=3in 
   \epsfbox{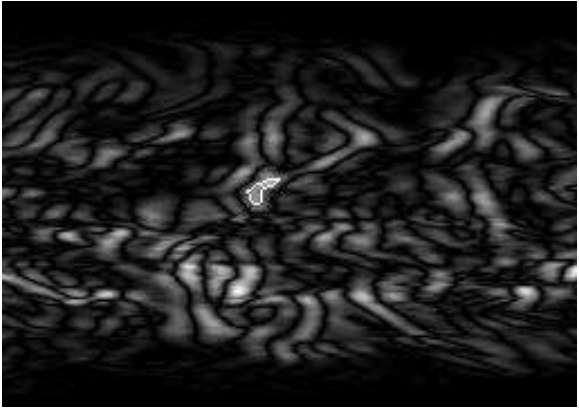}

	\end{center}
\caption{ Minimum variance connectedness determination without smoothing }
\end{figure}

When we smoothed the image, we got  $\lambda_v$=0.90, and the result turned to be correct, 
see Fig. 13.

\begin{figure}[hbt]
	\begin{center}

   \epsfxsize=3in 
   \epsfbox{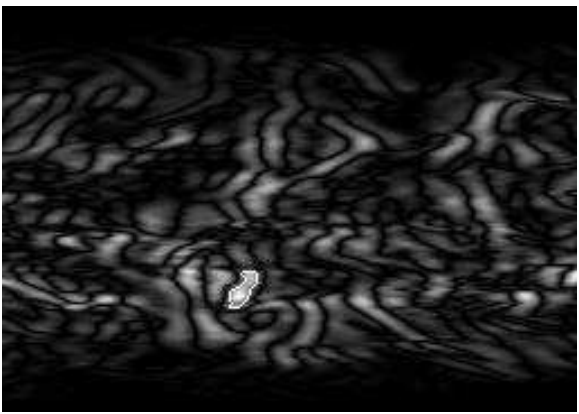}

	\end{center}
\caption{ Minimum variance connectedness determination with smoothing }
\end{figure}

\subsection{ $\lambda$-Connectedness and  Mumford-Shah's Method } 

How do we use Mumford-Shah's idea to find the optimal segmentation? We can define 
$L$ as the total length of the edge of all segments. 

\begin {equation}   
     H (\lambda ) = \alpha \cdot \Sigma \mbox{(inner variance of each $\lambda$-component)} + \beta \cdot L 	              
\end {equation} 

\noindent where $\alpha$ and $\beta$ are constants. More generally, we can do the 
normal $\lambda$-connected fit \cite{Che91} \cite{CA} on each $\lambda$-connected
component. The total variance (or standard deviation) of the (normal $\lambda$-connected) 
fitted image is denoted as $V$. $L$ is still the
total length of edges of segments (components), and $D$ is the difference between the fitted
image and the original image. Using  Mumford-Shah's Method, we can minimize the following
equation to get the $\lambda_p$ .   

\begin {equation}   
     H (\lambda ) = \alpha \cdot (V) + \beta \cdot (L) + \gamma \cdot (D)	              
\end {equation}

\begin {equation} 
    H (\lambda_p ) = \min \{ H (\lambda ) |    \lambda \in [0,1]\}
\end {equation} 

\vskip 10pt

\section {Discussion}
  
  Even though we calculated the entropy or variance in each connected component that is different from 
  the standard maximum entropy and the Otsu's method in image segmentation, the philosophy
  remains the same as in these two popular methods. The results are very promising.
  These two new methods can be easily applied in other region-growing segmentations. A large amount
  of further research should be done to support and the new methods. We will implement  the method proposed 
  in subsection E in section III, and compare it with the results obtained in \cite{CV}.


\begin{thebibliography}{99}
%

\bibitem{STI} Abdulkadir Sengur, Ibrahim Turkoglu, M. Cevdet Ince, 
A comparative  study on entropic thresholding methods, 
Istanbul University Journal of Electrical $\&$ Electronics Engineering Year
Vol 6 No 2, pp 183-188, 2006.

 %
\bibitem{CV} T. Chan and L. Vese, Active contours without edges.
IEEE Trans on Image Processing, Vol 10, No 2 pp 266-277.
%

\bibitem{Che85} L. Chen Three-dimensional fuzzy digital topology and its
        applications(I), {\it Geophysical Prospecting for petroleum}, Vol 24, No 2,
        pp 86-89, 1985.
%
\bibitem{Che91} L. Chen, The lambda-connected segmentation and the optimal algorithm
        for split-and-merge segmentation,  {\it Chinese J. Computers}
        Vol 14, pp 321-331, 1991.
%

\bibitem{Che02} L. Chen, $\lambda$-connected approximations for rough sets, In Lecture Notes in Computer Science, 
Springer, Vol 2457, 572-577, 2002.
%
\bibitem{Che04} L. Chen, Discrete Surfaces and Manifolds: A theory of digital-discrete geometry and topology, 
           S$\&$P Computing, 2004
%

\bibitem{Che06}	L. Chen, $\lambda$-Measure for Bone Density Connectivity, Proceedings 
of IEEE International Symposium on Industrial Electronics, 2006 Montreal, Quebec, Canada, 489-494.

%
\bibitem{CCZ} L. Chen, H.D. Cheng,  and J. Zhang,  Fuzzy subfiber and its application
         to seismic lithology classification, {\it Information Science: Applications},
         Vol 1, No 2, pp 77-95, 1994.
%
\bibitem{CA} L. Chen, and O. Adjei  $\lambda$-Connected Segmentation and Fitting:
Three New Algorithms, Proceedings of IEEE conference on System, Man, and Cybernetics 2004.

%

\bibitem{CAC}L. Chen, O. Adjei, and D. H. Cooley, $\lambda$-Connectedness : Method
and Application, {\it Proceedings of IEEE conference on System, Man, and
Cybernetics 2000,} pp 1157-1562, 2000.
%

\bibitem{CV} M. Cheriet, J.N. Said, C.Y. Suen, A recursive thresholding technique for 
  image segmentation, {\it IEEE Transection on Iamge Processing}, vol 7 No 6, 1998, 918-921

\bibitem{CLR}T. H. Cormen, C.E. Leiserson,  and R.L. Rivest, {\it Introduction to Algorithms},
        MIT Press, 1993.
%
\bibitem{GW} R. C. Gonzalez, and R. Wood, {\it Digital Image Processing},
        Addison-Wesley, Reading, MA, 1993.
%
\bibitem{HS} L. Hertz and R. W. Schafer, Multilevel thresholding using edge
matching, {\it Comput. Vis. Graph. Image Process.}, vol. 44, pp. 279-295,
1988.

\bibitem{KSW} Kapur J.N., Sahoo P.K. and Wong A.K.C., A new method of gray level 
picture thresholding using the entropy of the histogram, {\it 
Comput. Vision Graphics Image Process.},29, 273-285, 1985.

\bibitem{Mount}T. Kanungo, D. M. Mount, N. Netanyahu, C. Piatko, R. Silverman, and 
A. Y. Wu,An efficient k-means clustering algorithm: Analysis and
implementation, {\it IEEE Trans. Pattern Analysis and Machine Intelligence}, 24 (2002), 881-892.
 
\bibitem{KI} J. Kittler and J. Illingworth, Minimum error thresholding, {\it Pattern
Recognit.}, vol. 19, pp. 41-47, 1986.
%
\bibitem{Koh}  R. Kohler, A segmentation system based on thresholding, {\it Comput.
Graphics Image Process.}, vol. 15, pp. 319-338, 1981.
 
\bibitem{LKZC} C-T Lu, Y. Kou, J. Zhao, and L. Chen, Detecting and tracking region 
            outliers in meteorological data, {\it Information Sciences}, 2007, pp 1609-1632.

\bibitem{MS} D. Mumford and J. Shah, Optimal approximation by piecewise smooth
functions and associated variational problems, {\it Communication of Pure and Applied
Mathematics}, vol. 42, pp. 577-685, 1989.

\bibitem{Ots} N. Otsu, A threshold selection method from grey-level histograms,
{\it IEEE Trans. Syst., Man, Cybern.}, vol. SMC-8, pp. 62-66, 1978.

\bibitem{PP} N. Pal and S. Pal, A review on image segmentation techniques, {\it Pattern Recognition}, Vol 26,
pp 1277-1294, 1993 .

\bibitem{Pav} T. Pavilidis, Algorithms for Graphics and Image Processing, {\it
Computer Science Press}, Rockville, MD, 1982.

 \bibitem{Hal} T. A. Runkler, J. C. Bezdek, and L. O. Hall,
Clustering very large data sets: the complexity of the fuzzy c-means
algorithm, {\it Proc. EUNITE 2002, ed. K. Lieven, publ. By Elite Fndn,
Aachen}, Germany, ISBN 3-89653-919-1, 420-425, 2002.
%
\bibitem{SSW}  P. K. Sahoo, S. Soltani, and A. K. C. Wong, SURVEY: A survey of
thresholding techniques, {\it Comput. Vis. Graph. Image Process.}, vol. 41,
pp. 233-260, 1988.
%
\bibitem{SW} M. Spann and R. Wilson, A quad-tree approach to image segmentation
which combines statistical and spatial information, {\it  Pattern Recognit.},
vol. 18, pp. 257-269, 1985.

\bibitem{TB} L. Tsai and F.T. Berkey, Ionogram analysis using fuzzy segmentation and connectedness techniques,
{\it Radio Science}, Vol 35, No 2, 1173-1186, 2000.
%
\bibitem{TBL} L. Tsai, F. T. Berkey, and J. Y. Liu,
Automatic ionogram trace identification using fuzzy classification techniques,
{\it Computer Aided Processing of Ionograms and Ionosonde Records}, France, pp 45-50, 1996.
%
\bibitem{WH} S. Wang and R. M. Haralick, Automatic multithreshold selection,
{\it Comput. Vis. Graph. Image Process.}, vol. 25, pp. 46-67, 1984.


\end{thebibliography}
\end{document}